\documentclass[conference]{IEEEtran}
\IEEEoverridecommandlockouts

\usepackage[utf8]{inputenc}
\usepackage{todonotes}
\usepackage{amsmath}
\usepackage{amssymb}
\usepackage{amsthm}
\newtheorem{theorem}{Theorem}
\usepackage{bbm}
\usepackage[noend]{algpseudocode}
\usepackage{algorithm}
\usepackage{booktabs}
\usepackage{multirow}
\usepackage{subcaption}
\usepackage{caption}
\usepackage{graphicx}
\usepackage{xcolor}
\usepackage{pgfplots}
\usepackage{caption}
\usepackage{subcaption}
\usepackage[bottom]{footmisc}
\pgfplotsset{compat=newest}
\definecolor{1F77B4}{HTML}{1F77B4}
\definecolor{FF7F0E}{HTML}{FF7F0E}
\definecolor{2CA02C}{HTML}{2CA02C}
\definecolor{D62728}{HTML}{D62728}
\definecolor{9467BD}{HTML}{9467BD}
\definecolor{8C564B}{HTML}{8C564B}
\definecolor{E377C2}{HTML}{E377C2}
\definecolor{7F7F7F}{HTML}{7F7F7F}
\definecolor{BCBD22}{HTML}{BCBD22}
\definecolor{17BECF}{HTML}{17BECF}
\def\BibTeX{{\rm B\kern-.05em{\sc i\kern-.025em b}\kern-.08em
    T\kern-.1667em\lower.7ex\hbox{E}\kern-.125emX}}

\begin{document}

\title{Towards Explainable Bit Error Tolerance of Resistive RAM-Based Binarized Neural Networks}


\author{

\IEEEauthorblockN{
Sebastian Buschjäger\IEEEauthorrefmark{1}\IEEEauthorrefmark{3},
Jian-Jia Chen\IEEEauthorrefmark{2},
Kuan-Hsun Chen\IEEEauthorrefmark{2}, 
Mario Günzel\IEEEauthorrefmark{2}\IEEEauthorrefmark{3}, 
Christian Hakert\IEEEauthorrefmark{2},\\
Katharina Morik\IEEEauthorrefmark{1},
Rodion Novkin\IEEEauthorrefmark{2},
Lukas Pfahler\IEEEauthorrefmark{1}\IEEEauthorrefmark{3},
and Mikail Yayla\IEEEauthorrefmark{2}\IEEEauthorrefmark{3}
}

\IEEEauthorblockA{
\textit{\IEEEauthorrefmark{1}Artificial Intelligence Group}, \textit{TU Dortmund University, Germany}
}

\IEEEauthorblockA{
\textit{\IEEEauthorrefmark{2}Design Automation for Embedded Systems Group}, \textit{TU Dortmund University, Germany} 
}

\IEEEauthorblockA{
\textit{\IEEEauthorrefmark{3}These authors contributed equally to this research}
}

\IEEEauthorblockA{\{sebastian.buschjaeger, jian-jia.chen, kuan-hsun.chen, mario.guenzel, christian.hakert,\\ katharina.morik, rodion.novkin, lukas.pfahler, mikail.yayla\}@tu-dortmund.de}
{\footnotesize 
\thanks{This paper has been supported by Deutsche Forschungsgemeinschaft (DFG), as part of the Collaborative Research Center SFB 876 ”Providing Information by Resource-Constrained Analysis”, project A1 (http://sfb876.tu-dortmund.de) and Project OneMemory (CH 985/13-1).}
}

}



\maketitle
\begin{abstract}
Non-volatile memory, such as resistive RAM (RRAM), is an emerging energy-efficient storage, especially for low-power machine learning models on the edge.
It is reported, however, that the bit error rate of RRAMs can be up to 3.3\% in the ultra low-power setting, which might be crucial for many use cases.
Binary neural networks (BNNs), a resource efficient variant of neural networks (NNs), can tolerate a certain percentage of errors without a loss in accuracy and demand lower resources in computation and storage.
The bit error tolerance (BET) in BNNs can be achieved by flipping the weight signs during training, as proposed by Hirtzlin et al., but their method has a significant drawback, especially for fully connected neural networks (FCNN): The FCNNs overfit to the error rate used in training, which leads to low accuracy under lower error rates.
In addition, the underlying principles of BET are not investigated.
In this work, we improve the training for BET of BNNs and aim to explain this property.
We propose straight-through gradient approximation to improve the weight-sign-flip training, by which BNNs adapt less to the bit error rates.
To explain the achieved robustness, we define a metric that aims to measure BET without fault injection.
We  evaluate  the  metric  and  find  that  it  correlates with accuracy over error rate for all FCNNs tested.
Finally, we explore the influence of a novel regularizer that optimizes with respect to this metric, with the aim of providing a configurable trade-off in accuracy and BET. 


\end{abstract}




\section{Introduction}

In the age of ubiquitous computing, sensors and computing facilities are embedded into various physical environments for data collection. Small devices apply machine learning models on data streams directly on the edge.
Since the edge devices have resource constraints,
such as in computation and storage, these models need to be efficient in execution and memory usage.
Binary neural networks (BNNs) are one resource-efficient variant of neural networks (NNs), which are especially well suited for small embedded devices.
Their weight parameters are stored as binary values, and the convolution operations are computed with XNOR followed by population count (POPCOUNT) instructions, which count the number of set bits.
The trade-off for resource-efficiency is a decrease in accuracy by a few percentage points compared to full-precision neural networks.
The efficient execution of BNNs has been researched in several recent works~\cite{DBLP:journals/corr/CourbariauxB16, DBLP:journals/corr/abs-1811-11431,Yang:2017:BOB:3123266.3129393}, but the memory type to use for BNN models in a low-power setting has received limited attention so far, despite the energy saving potential.

Non-volatile memories (NVMs), such as resistive random-access memory (RRAM), are emerging memory technologies for low-power storage. 
They are expected to be deployed in future computing systems with resource constraints \cite{boukhobz/etal/2017}.
Because of their non-volatility, they make normally-off computing efficient: The device is only powered on if there is computation to be done.
This is especially convenient for inference on the low-power edge.
RRAM, which stores information in the form of non-volatile resistive states, has comparable performance to DRAM, but uses less energy because no refreshs are needed and the read/write energy is lower \cite{boukhobz/etal/2017}. 
Moreover, when using an ultra low-power setting for RRAM cell programming, the energy consumption can be lowered further, up to 30 times for the programming energy, as reported by Hirtzlin et al. \cite{hirtzlin/etal/2019}. 
This addresses one the major disadvantages of RRAMs: Cell lifetime. The lower programming energy stresses the cells less, which leads to increased lifetime.

The crucial drawback of the ultra low-power RRAM setting is, however, the high bit error rate of $\sim{3.3\%}$.
Hirtzlin et al. propose to use this setting for in-memory processing of BNNs, i.e. executing the BNN operations inside the memory, and show that BNNs can be trained to be more error tolerant, up to a rate of 4\% without a significant accuracy drop.
The increased bit error tolerance (BET) lowers the requirements on the memory and makes possible the use of the ultra low-power setting for BNNs.
RRAM is therefore a highly promising low-power memory for BNNs.

The method proposed by Hirtzlin et al. \cite{hirtzlin/etal/2019} is simple: During training a certain percentage of bit errors are introduced into the network by randomly flipping weights. Even though this method is very effective, it has several drawbacks.
First, the accuracy drop is considerable, especially for fully connected neural networks (FCNN).
In their experiments, the FCNNs adapt to the error rates they were trained for, which means that the accuracy of the BNNs drop in cases in which a lower percentage of errors is present.
Secondly, faults have to be injected during training, which adds complexity to the training process. 
As discussed in \cite{8013784}, training NNs for general BET (i.e. without accuracy drops for lower error rates) is not a trivial task.
The explanation of the underlying principles of the BET of BNNs is not explicitly studied in the literature as well.


In this work, we report on our progress and future directions for \textit{achieving general bit error tolerance} and for \textit{explaining} this property:
\begin{itemize}
  \item We improve the BET training of the previous work by using straight-through gradient approximation, so that the NNs do not overfit to the error rates used during training.
  \item We present a metric that aims to measure the achieved BET of BNNs, without injecting faults.
  \item Based on this metric, we explore the impact of a regularizer with the goal of achieving general BET, which is a property independent from the error model.
\end{itemize}

The paper is organized as the following.
Section \ref{sec:BNNs} introduces BNNs formally.
Section~\ref{sec:RobustnessBNNs} formalizes the BET of NNs
, whereas Section~\ref{sec:TrainingRobustBNNs} presents a novel regularizer to enhance the BET of NNs.
Section~\ref{sec:RelatedWork} surveys the related work, whereas Section~\ref{sec:Experiments} presents our experiments.
Section \ref{sec:Conclusion} concludes the paper.

\section{Binarized Neural Networks}
\label{sec:BNNs}


To train a binarized neural network with weights in $\mathbb{F}_2 = \{-1,+1\}$ we follow the approach presented by Hubara et al. \cite{Hubara/etal/2016} in which the weights are stored as floating point numbers, but both weights and activations are deterministically rounded to $\mathbb{F}_2$ during forward computation. The gradient updates are performed with full precision on the floating point weights.

\subsection{Notation}

Before we describe the training procedure in more detail, we introduce the notation used to describe neural networks. We assume a feed-forward network in which each layer $l$ is associated with a weight tensor $W^l$. Each layer performs a generic operation $\circ$ to compute its output $h^l(X) := W^l \circ X $ given its input tensor $X^{l-1}$ and weights in $W^l$.
For example, a fully-connected layer computes the matrix product $h^l(X^{l-1}) = W^l X^{l-1}$ or the convolution layer computes a convolution with a number of filters defined by $W$ denoted by $h^l(X^{l-1}) = W^l \ast X^{l-1}$. Between layers we apply an activation function $\sigma(h^l(X^l))$ in order to obtain a non-linear decision function. 
In BNNs it is common to use the sign function as activation function.



\subsection{Training}

Floating point networks are typically trained with gradient-based approaches such as mini-batched stochastic gradient descent (SGD) to minimize a loss function. Let
$\mathcal D = \{(x_1,y_1),\dots,(x_I,y_I)\}$ with $x_i \in \mathcal X$ and $y_i \in \mathcal Y$
 denote the training data and let $\ell \colon \mathcal Y \times \mathcal Y \to \mathbb R$ be the loss function.
Let $W = (W^1,\dots, W^L)$ denote the weight tensors of each layer in the neural network and let $f_W(x)$ be the output of the network given its weights $W$, then we aim to solve the following optimization problem
$$
\arg\min_{W} \frac{1}{I}\sum_{(x,y)\in\mathcal D} \ell(f_{W}(x), y)
$$
by a gradient descent strategy that computes the gradient $\nabla_W \ell$ using backpropagation.
Unfortunately, in the case of binary neural networks we cannot perform gradient-based optimization directly. This is due to two reasons: First, the space of weights is discrete and thus the parameter-vector obtained by taking a small step in the opposite direction of the gradient is almost certainly not binary. Second, the sign-function is not differentiable and also its sub-differential is useless for optimization, as it is zero everywhere other than zero.

To mitigate the first problem, Hubara et al. \cite{Hubara/etal/2016} propose a scheme that during training stores weights as floating point numbers constrained to values between -1 and 1 and `binarizes' the network during the forward pass.
\begin{algorithm}[t]
  \begin{algorithmic}[1]
    \Function{forward}{model, $x$}
      \For{$l \in \{1,\dots,L\}$}
        \State{$x \gets B(B(W^l) \circ x)$}
      \EndFor
      \State \Return $x$
    \EndFunction
  \end{algorithmic}
  \caption{Binarized forward pass $f_W(x)$.}
  \label{fig:Forward}
\end{algorithm}
More formally, let $b \colon \mathbb R \to \mathbb{F}_2$ be a binarization function with
$$
b(x) = \begin{cases}
          1 & x > 0 \\
          -1 & \, \text{else}
       \end{cases}
$$
and let $B(W^l)$ denote the element-wise application of $b$ to $W^l$. 
We summarize the forward-pass in Algorithm~\ref{fig:Forward}.
The outer application of $B$ in Algorithm~\ref{fig:Forward} acts as the activation function and that $b$ can also be interpreted as an un-smooth version of the $tanh$ activation function. Therefore it is sometimes called hard-tanh or $Htanh$.

Then, during the backward pass they use full floating point precision. 
To mitigate the second problem -- $b$ is not differentiable -- they replace the gradient of $b$ with the so-called straight-through estimator. Consider the forward computation $Y = B(X)$. Let $\nabla_Y \ell$ denote the gradient with respect to $Y$. The straight-through estimator approximates
\begin{equation}
\nabla_X \ell := \nabla_Y \ell,\label{eq:straight-through}
\end{equation} essentially pretending that $B$ is the identity function. Using this gradient approximation we can apply standard stochastic gradient descent techniques with the small addition that all floating point weights are clipped to be between -1 and 1 after each gradient update.

For faster and more reliable training, we use the standard deep learning technique Batch Normalization. We insert batch normalization layers between layers and the following activation functions. The batch normalization layers shift and scale the outputs computed by the respective layers, then the sign function is applied. Hence the forward pass can still be computed using only binary arithmetics, the batch normalization just shifts the threshold of the binary activation from zero to a data-dependent number. One peculiarity of our models is that we apply normalization also after the last linear layer, before the outputs are fed into a softmax-layer with subsequent cross-entropy loss. While this seems counter-intuitive at first, we find that it improves loss and eases training substantially. We suspect that this is due to the rescaling of outputs: Plain binary networks output large integer activations on the last layer which, fed into softmax activations, often result in vanishing gradients.

\section{Bit Error Tolerance of BNNs}
\label{sec:RobustnessBNNs}

To understand the error tolerance of BNNs we propose to formalize it using a metric that is calculated on the neuron level with only one pass over the evaluation set. To do so we focus our efforts on CNNs. Please note that all of our definition are also applicable for FCNN if we view their inputs as $1\times1$ images. 
We first define the local error tolerance of a $2d$ feature map in a CNN. Then we leverage this definition into the error tolerance of a single neuron, which finally enables us to define the error tolerance of the whole 
network. 

Consider a CNN and let $n$ be the index of one neuron. Recall that the output of a neuron is a $2d$ feature map with height $U$ and width $V$.
%
We define the neuron's \emph{local error tolerance} $T_{i,n,u,v}$ at position $u,v$ by modeling the number of weight sign flips it can tolerate without a change of its output given the input $x_i$.
To do so, let $h_{i,n,u,v}$ be the output of $n$-th neuron \emph{before} applying the activation function.
For neurons which are not in the first layer, we note two things:
First, each neuron's output is computed by a weighted sum of inputs that are $\pm 1$ with weights that are also $\pm 1$.
Second, the sign function is applied to this output.
Thus as long as weight flips do not change the sign of the weighted sum, a neuron is error tolerant.
Formally, we can quantify the error tolerance of a neuron given the input $x_i$ by the distance of its output from $0$: 
\begin{equation} \label{eq:Tofy}
  T_{i,n,u,v} = \left|h_{i,n,u,v} - s_n -\tfrac{1}{2}\right|.
\end{equation}
We include $s_n$ to account for activation shifts due to the Batch Normalization Layer (without BatchNorm $s_n=0$), and to avoid ambiguity at $0$ we subtract $\frac{1}{2}$.
Note that $T_{i,n,u,v}$ is a measure for the worst case error tolerance, in the sense that at least $\lfloor \frac{T_{i,n,u,v}}{2} \rfloor + 1$ weight sign flips, are necessary. 
With each weight sign flip $h_{i,n,u,v}$ can get closer to $s_n$ and finally flip the output. 

The definition of $T_{i,n,u,v}$ yields the following theorem:

\begin{theorem}\label{thm:tolerateflip}
	Let $b \in \mathbb{R}_{\geq 0}$.
	If $T_{i,n,u,v} \geq b$ for all $u,v$ then the neuron can tolerate at least $\lfloor \frac{b}{2} \rfloor$ bitflips, i.e. any bitflip of $\lfloor \frac{b}{2} \rfloor$ weights of the neuron does not affect its output.
\end{theorem}

The proof can be found in the appendix.




Intuitively, a neuron is error tolerant if it is robust across all positions. Thus, we may demand that each position has a local error tolerance of at least $b$. More formally, the \emph{error tolerance} $T^{b}_{i,n}$ of a neuron $n$ given the input $x_i$ is defined as:

\begin{equation} \label{eq:Tn}
  T^{b}_{i,n} = \frac{1}{UV} \sum_{u=1}^{U} \sum_{v=1}^{V} \mathbbm{1} \{ T_{i,n,u,v} \geq b \}.
\end{equation}

The error tolerance of the whole network can then be defined as the average error tolerance across all neurons:
\begin{equation} \label{eq:TIB}
   T_i^{b} = \frac 1 {N} \sum_{n=1}^{N} T^{b}_{i,n}.
\end{equation}

We determine $T^{b}$ for a network by evaluating the BNN on the full data set:
\begin{equation} \label{eq:TB}
   T^{b} = \frac 1 I \sum_{i=1}^{I} T^{b}_{i}.
\end{equation}

For neurons in the first layer, we assume that the inputs are not in $\{\pm1\}$ but $\{0,\dots,Z\}$.
Thus we have to scale the local error tolerance:
\begin{equation} \label{eq:Tofyfirst}
T_{i,n,u,v} = \frac{\left|h_{i,n,u,v} - s_n -\tfrac{1}{2}\right|}{Z}.
\end{equation}

With the definition of $T^b$, we aim to explain the robustness of BNNs against bit errors without fault injection.

\section{Training Bit Error Tolerant Neural Networks}
\label{sec:TrainingRobustBNNs}

In this section we propose two different ways to regularize the BNN training objective to account for bit errors during training and achieve bit error tolerance.
The first approach is based directly on the insights in Section~\ref{sec:RobustnessBNNs}, and the second one is based on flip-training as proposed by Hirtzlin et al. \cite{hirtzlin/etal/2019}.
\subsection{Direct Regularization}
\label{subsec:directreg}
As discussed in Section~\ref{sec:RobustnessBNNs}, a high $T^b$-value for a neuron indicates that many weight signs can flip without changing the activation of the neuron. The quantity $T^b$ is not a differentiable function but essentially a count. However, we can still construct a regularizer that punishes those neurons that do not have a flip-tolerance of at least $b$.
We rely on the well-known hinge function to build a convex and sub-differentiable regularizer.
For a given bit-flip tolerance level $b$, we propose to regularize each neuron $n$ for each input example $i$ using the hinge-function
\begin{equation}
R_{n,u,v}^b(x_i)=\max(0,b - T_{i,n,u,v}).
\end{equation}
Whenever a neuron has a $T_{i,n,u,v}$-value of at least $b$, the minimum of $R$ is achieved.
To regularize the whole network, we compute the mean of all neuron regularizers. We weight the regularizer with $\lambda>0$ and add it to the loss.
\subsection{Flip Regularization}
\label{subsec:flipreg}
Flip regularization is a technique first proposed by Hirtzlin et al. \cite{hirtzlin/etal/2019}. The idea is simple: To make the network robust against bit errors, we simulate those errors already during training time. During each forward-pass computation, we generate a random bitflip-mask and apply it to the binary weights.
However, there are two ways to implement this.
Let $M$ denote a random bitflip mask with entries $\pm 1$ of the same size as $W$ that we multiply component-wise to the binarized weights.
We first consider computing the bit-flip operation as $H = (B(W) \cdot M)\circ X$. Standard backpropagation on a loss $\ell$ that is a function of $H$ yields the following gradient of $\ell$ with respect to $B(W)$
$$\nabla_{B(W)} \ell= M \cdot \nabla_{B(W) \cdot M} \ell$$
which e.g. for fully connected layers amounts to a gradient update
$$\nabla_{B(W)} \ell = M \cdot (\nabla_H\ell\; X^T).$$
We see that an update computed this way is aware of the bit-flips that were performed and accounts for them. We propose instead to use a special flip-operator with straight-through gradient approximation.
We denote by $e_p$ the bit error function that flips its input with probability $p$ and let $E_p$ denote its component-wise counterpart. During training we change the forward pass such that it computes
$$X^{l+1} := B(E_p(B(W^l)) \circ X^l).$$
We replace the gradient of $E_p$ with a straight-through approximation as in (\ref{eq:straight-through}). This way, in the example above we now have $H = E_p(B(W)) \circ X$ with gradient updates $\nabla_{B(W)} \ell = \nabla_{E_p(B(W))} \ell$ which for fully connected layers yields the update
$$\nabla_{B(W)} \ell = \nabla_H \ell\;X^T$$
which is unaware of bit-flips and just uses the corrupted outputs $H$.

We believe that the approach using straight-through gradient approximation is superior and that the problems reported by Hirtzlin et al. \cite{hirtzlin/etal/2019} can be sourced to them using the native implementation. Particularly as we will see in Section~\ref{sec:Experiments}, our implementation does not overfit to a particular error probability.

\section{Experiments}
\label{sec:Experiments}

\begin{table}[] 
  \center
  \begin{tabular}{@{}llrrr@{}}
  \toprule
  Name         & \# Train & \# Test & \# Dim & \# classes       \\ \midrule
  FashionMNIST\footnote{{https://github.com/zalandoresearch/fashion-mnist}} & 60000    & 10000   & (1,28,28)  & 10   \\
  CIFAR10     & 50000    & 10000   & (3,32,32)   & 10  \\ \bottomrule
  \end{tabular}
  \caption{Datasets used for experiments.}
  \label{tab:datasets}
\end{table}


\begin{table}[]
  \center
  \begin{tabular}{@{}ll@{}}
  \toprule
  Parameter                                   & Range                                                        \\ \midrule
  Regularization                              & $\lambda \in \{10^{-4}, 10^{-3}, 10^{-2}\}$                  \\
  Flip probability                            & $p \in \{0.01, 0.05, 0.1, 0.2\}$                             \\
  Robustness                                  & $b \in \{32,64,128\}$                                        \\[1mm]
  Fashion FCNN & In $\to$ FC 2048 $\to$ FC 2048 $\to$ 10                \\
  Fashion CNN & In $\to$ C64 $\to$ MP 2 $\to$ C64 $\to$ MP 2 \\
  & \phantom{In} $\to$ FC2048 $\to$ FC2048 $\to$ 10 \\[1mm]
  CIFAR10 CNN & In $\to$ C128 $\to$ C128 $\to$ MP 2 $\to$ C256 $\to$ C256 \\
  & \phantom{In} $\to$ MP 2 $\to$ C256 $\to$ C256 $\to$ MP 2 \\
  & \phantom{In} $\to$ FC 2048 $\to$ FC 2048 $\to$ 10 \\ \bottomrule
  \end{tabular}
  \caption{Parameters used for experiments.}
  \label{tab:expsetting}
\end{table}

\begin{figure*}[!htb]
    \begin{subfigure}[b]{0.32\textwidth}
     %


\begin{tikzpicture}
	\begin{axis}[
		width=\linewidth,
		width=6cm,
    height=5cm,
		xlabel=Bit Error Rate (\%),
		ytick distance=1,
		ylabel=Accuracy (\%),
		legend pos=south west,
		title = \textbf{FCNN FashionMNIST}
	]

		\addplot+[line width=1pt,mark=none,color=1F77B4,error bars/.cd,y dir=both, y explicit] coordinates { (0.0, 90.04) +- (0.04999999999999716,0.04999999999999716)
			(0.5, 89.795) +- (0.09499999999999886,0.09499999999999886)
			(1.0, 89.77000000000001) +- (0.240000000000002,0.240000000000002)
			(1.5, 89.595) +- (0.09499999999999886,0.09499999999999886)
			(2.0, 89.455) +- (0.045000000000001705,0.045000000000001705)
			(2.5, 88.84) +- (0.21999999999999176,0.21999999999999176)
			(3.0, 88.91) +- (0.01999999999999602,0.01999999999999602)
			(3.5, 88.70500000000001) +- (0.02499999999999858,0.02499999999999858)
			(4.0, 88.41499999999999) +- (0.10499999999999687,0.10499999999999687)
			(4.5, 88.095) +- (0.015000000000000568,0.015000000000000568)
			(5.0, 87.79499999999999) +- (0.15499999999999403,0.15499999999999403)
			(5.5, 87.195) +- (0.005000000000002558,0.005000000000002558)
			(6.0, 86.805) +- (0.09500000000000597,0.09500000000000597)
			(6.5, 86.03) +- (0.020000000000010232,0.020000000000010232)
			(7.0, 85.325) +- (0.25499999999999545,0.25499999999999545)
			(7.5, 85.47999999999999) +- (0.19000000000000483,0.19000000000000483)
			(8.0, 84.45500000000001) +- (0.17500000000000426,0.17500000000000426)
			(8.5, 84.015) +- (0.14499999999999602,0.14499999999999602)
			(9.0, 82.905) +- (0.2749999999999915,0.2749999999999915)
			(9.5, 82.33) +- (0.0,0.0)
			(10.0, 81.41) +- (0.11000000000000654,0.11000000000000654) };
		\addlegendentry{No Reg.}
		\addplot+[line width=1pt,mark=none,color=FF7F0E,error bars/.cd,y dir=both, y explicit] coordinates { (0.0, 90.07) +- (0.07000000000000028,0.07000000000000028)
			(0.5, 89.86500000000001) +- (0.015000000000007674,0.015000000000007674)
			(1.0, 89.82499999999999) +- (0.10499999999999687,0.10499999999999687)
			(1.5, 89.905) +- (0.11499999999999488,0.11499999999999488)
			(2.0, 89.6) +- (0.14999999999999858,0.14999999999999858)
			(2.5, 89.515) +- (0.3149999999999977,0.3149999999999977)
			(3.0, 89.6) +- (0.020000000000003126,0.020000000000003126)
			(3.5, 89.48) +- (0.010000000000005116,0.010000000000005116)
			(4.0, 89.295) +- (0.045000000000001705,0.045000000000001705)
			(4.5, 89.11) +- (0.010000000000005116,0.010000000000005116)
			(5.0, 89.155) +- (0.15500000000000114,0.15500000000000114)
			(5.5, 88.995) +- (0.15499999999999403,0.15499999999999403)
			(6.0, 88.575) +- (0.11499999999999488,0.11499999999999488)
			(6.5, 88.33) +- (0.0,0.0)
			(7.0, 88.235) +- (0.06499999999999773,0.06499999999999773)
			(7.5, 87.625) +- (0.2149999999999963,0.2149999999999963)
			(8.0, 87.74000000000001) +- (0.020000000000003126,0.020000000000003126)
			(8.5, 87.305) +- (0.32499999999999574,0.32499999999999574)
			(9.0, 86.815) +- (0.11499999999999488,0.11499999999999488)
			(9.5, 86.65) +- (0.09999999999999432,0.09999999999999432)
			(10.0, 85.875) +- (0.34499999999999886,0.34499999999999886) };
		\addlegendentry{flip, $p=0.1$}
		\addplot+[line width=1pt,mark=none,color=2CA02C,error bars/.cd,y dir=both, y explicit] coordinates { (0.0, 89.15) +- (0.00999999999999801,0.00999999999999801)
			(0.5, 89.13) +- (0.05999999999999517,0.05999999999999517)
			(1.0, 89.275) +- (0.16499999999999915,0.16499999999999915)
			(1.5, 89.39500000000001) +- (0.09500000000000597,0.09500000000000597)
			(2.0, 89.13) +- (0.07000000000000739,0.07000000000000739)
			(2.5, 89.18) +- (0.07000000000000028,0.07000000000000028)
			(3.0, 89.16) +- (0.04999999999999716,0.04999999999999716)
			(3.5, 89.04) +- (0.25,0.25)
			(4.0, 89.13999999999999) +- (0.0800000000000054,0.0800000000000054)
			(4.5, 88.92) +- (0.01999999999999602,0.01999999999999602)
			(5.0, 88.95) +- (0.06999999999999318,0.06999999999999318)
			(5.5, 88.91) +- (0.00999999999999801,0.00999999999999801)
			(6.0, 88.93) +- (0.03999999999999915,0.03999999999999915)
			(6.5, 88.61) +- (0.0799999999999983,0.0799999999999983)
			(7.0, 88.6) +- (0.0899999999999963,0.0899999999999963)
			(7.5, 88.44500000000001) +- (0.17499999999999716,0.17499999999999716)
			(8.0, 88.4) +- (0.11999999999999744,0.11999999999999744)
			(8.5, 87.965) +- (0.19500000000000028,0.19500000000000028)
			(9.0, 87.91) +- (0.23999999999999488,0.23999999999999488)
			(9.5, 87.78999999999999) +- (0.19000000000000483,0.19000000000000483)
			(10.0, 87.35) +- (0.14000000000000057,0.14000000000000057) };
		\addlegendentry{flip, $p=0.2$}

	\end{axis}
\end{tikzpicture}
    \end{subfigure}%
    \begin{subfigure}[b]{0.32\textwidth}
     \begin{tikzpicture}
	\begin{axis}[
		xlabel=Bit Error Rate (\%),
    width=6cm,
    height=5cm,
    legend pos=south west,
		title = \textbf{CNN FashionMNIST},
	]
		\addplot+[line width=1pt, mark=none,color=1F77B4,error bars/.cd,y dir=both, y explicit] coordinates { (0.0, 91.17500000000001) +- (0.09499999999999176,0.09499999999999176)
			(0.5, 90.92500000000001) +- (0.005000000000002558,0.005000000000002558)
			(1.0, 90.75) +- (0.25,0.25)
			(1.5, 90.555) +- (0.19500000000000028,0.19500000000000028)
			(2.0, 90.38) +- (0.4899999999999949,0.4899999999999949)
			(2.5, 90.185) +- (0.10500000000000398,0.10500000000000398)
			(3.0, 89.655) +- (0.005000000000002558,0.005000000000002558)
			(3.5, 89.265) +- (0.19499999999999318,0.19499999999999318)
			(4.0, 88.72999999999999) +- (0.21000000000000085,0.21000000000000085)
			(4.5, 88.26499999999999) +- (0.384999999999998,0.384999999999998)
			(5.0, 87.80000000000001) +- (0.6399999999999935,0.6399999999999935)
			(5.5, 87.045) +- (0.34499999999999886,0.34499999999999886)
			(6.0, 86.18) +- (0.38000000000000256,0.38000000000000256)
			(6.5, 84.805) +- (0.6749999999999972,0.6749999999999972)
			(7.0, 84.82) +- (0.7899999999999991,0.7899999999999991)
			(7.5, 83.35) +- (0.8000000000000043,0.8000000000000043)
			(8.0, 82.155) +- (1.1550000000000011,1.1550000000000011)
			(8.5, 80.47500000000001) +- (1.0450000000000017,1.0450000000000017)
			(9.0, 78.975) +- (0.3749999999999929,0.3749999999999929)
			(9.5, 76.81) +- (1.9000000000000057,1.9000000000000057)
			(10.0, 75.64000000000001) +- (2.0700000000000003,2.0700000000000003) };
		\addlegendentry{No Reg.}
		\addplot+[line width=1pt, mark=none,color=FF7F0E,error bars/.cd,y dir=both, y explicit] coordinates { (0.0, 91.395) +- (0.15500000000000114,0.15500000000000114)
			(0.5, 91.35499999999999) +- (0.11500000000000199,0.11500000000000199)
			(1.0, 91.16499999999999) +- (0.16499999999999915,0.16499999999999915)
			(1.5, 90.975) +- (0.22500000000000142,0.22500000000000142)
			(2.0, 90.67) +- (0.1799999999999926,0.1799999999999926)
			(2.5, 90.36500000000001) +- (0.1250000000000071,0.1250000000000071)
			(3.0, 90.235) +- (0.0049999999999954525,0.0049999999999954525)
			(3.5, 89.905) +- (0.24499999999999744,0.24499999999999744)
			(4.0, 89.825) +- (0.3149999999999977,0.3149999999999977)
			(4.5, 89.16) +- (0.28999999999999915,0.28999999999999915)
			(5.0, 89.16499999999999) +- (0.2949999999999946,0.2949999999999946)
			(5.5, 88.45) +- (0.10999999999999943,0.10999999999999943)
			(6.0, 88.25999999999999) +- (0.17999999999999972,0.17999999999999972)
			(6.5, 87.91) +- (0.21000000000000085,0.21000000000000085)
			(7.0, 86.72) +- (0.5500000000000043,0.5500000000000043)
			(7.5, 86.27) +- (0.6700000000000017,0.6700000000000017)
			(8.0, 85.195) +- (1.3950000000000031,1.3950000000000031)
			(8.5, 84.1) +- (0.740000000000002,0.740000000000002)
			(9.0, 83.41499999999999) +- (0.5749999999999957,0.5749999999999957)
			(9.5, 82.14) +- (1.4000000000000057,1.4000000000000057)
			(10.0, 79.845) +- (2.7249999999999943,2.7249999999999943) };
		\addlegendentry{flip, $p=0.05$}
		\addplot+[line width=1pt, mark=none,color=2CA02C,error bars/.cd,y dir=both, y explicit] coordinates { (0.0, 90.68) +- (0.11999999999999744,0.11999999999999744)
			(0.5, 90.58) +- (0.04000000000000625,0.04000000000000625)
			(1.0, 90.4) +- (0.0800000000000054,0.0800000000000054)
			(1.5, 90.395) +- (0.07500000000000284,0.07500000000000284)
			(2.0, 90.36500000000001) +- (0.134999999999998,0.134999999999998)
			(2.5, 90.0) +- (0.01999999999999602,0.01999999999999602)
			(3.0, 89.82) +- (0.13999999999999346,0.13999999999999346)
			(3.5, 89.58500000000001) +- (0.16499999999999915,0.16499999999999915)
			(4.0, 89.22) +- (0.030000000000001137,0.030000000000001137)
			(4.5, 89.09) +- (0.00999999999999801,0.00999999999999801)
			(5.0, 88.475) +- (0.4450000000000003,0.4450000000000003)
			(5.5, 88.225) +- (0.17500000000000426,0.17500000000000426)
			(6.0, 87.725) +- (0.26499999999999346,0.26499999999999346)
			(6.5, 87.44999999999999) +- (0.03999999999999915,0.03999999999999915)
			(7.0, 86.58000000000001) +- (0.3200000000000003,0.3200000000000003)
			(7.5, 85.83500000000001) +- (0.8350000000000009,0.8350000000000009)
			(8.0, 84.95) +- (0.10999999999999943,0.10999999999999943)
			(8.5, 84.305) +- (0.7449999999999973,0.7449999999999973)
			(9.0, 83.35499999999999) +- (0.32499999999999574,0.32499999999999574)
			(9.5, 82.17) +- (0.21999999999999886,0.21999999999999886)
			(10.0, 80.785) +- (0.4450000000000003,0.4450000000000003) };
		\addlegendentry{flip, $p=0.1$}

	\end{axis}
\end{tikzpicture}
    \end{subfigure}%
    \hspace{-0.6cm}
    \begin{subfigure}[b]{0.32\textwidth}
%
%
%
\begin{tikzpicture}
	\begin{axis}[
		width=\linewidth,
    height=5cm,
    width=6cm,
		xlabel=Bit Error Rate (\%),
		ytick distance=10,
		legend pos=south west,
		title=\textbf{CNN CIFAR10}
	]

		\addplot+[line width=1pt,mark=none,color=1F77B4,error bars/.cd,y dir=both, y explicit] coordinates { (0.0, 79.86) +- (0.44392191505564643,0.44392191505564643)
			(0.5, 78.69333333333333) +- (0.31478387647541417,0.31478387647541417)
			(1.0, 78.07) +- (0.5975505557412372,0.5975505557412372)
			(1.5, 76.7) +- (0.599054811070463,0.599054811070463)
			(2.0, 75.27) +- (0.21118712081943047,0.21118712081943047)
			(2.5, 73.07) +- (0.8591080646034379,0.8591080646034379)
			(3.0, 71.13) +- (1.2758526560696581,1.2758526560696581)
			(3.5, 68.52) +- (1.0987568733194177,1.0987568733194177)
			(4.0, 66.57000000000001) +- (0.7647657593450817,0.7647657593450817)
			(4.5, 63.373333333333335) +- (1.0317083998021086,1.0317083998021086)
			(5.0, 60.196666666666665) +- (0.6367277457612648,0.6367277457612648)
			(5.5, 58.02333333333333) +- (0.9654129801397052,0.9654129801397052)
			(6.0, 55.31666666666666) +- (0.1596524001977053,0.1596524001977053)
			(6.5, 52.27) +- (0.6723590310739249,0.6723590310739249)
			(7.0, 49.14666666666667) +- (0.24390344173235926,0.24390344173235926)
			(7.5, 46.42666666666667) +- (0.1438363267359443,0.1438363267359443)
			(8.0, 42.07666666666667) +- (0.8204605752597491,0.8204605752597491)
			(8.5, 39.903333333333336) +- (0.691873463061624,0.691873463061624)
			(9.0, 37.79) +- (0.33674916480965295,0.33674916480965295)
			(9.5, 34.72) +- (0.5481483983983424,0.5481483983983424)
			(10.0, 32.18333333333334) +- (0.5230891149911495,0.5230891149911495) };
		\addlegendentry{No Reg.}
		\addplot+[line width=1pt,mark=none,color=FF7F0E,error bars/.cd,y dir=both, y explicit] coordinates { (0.0, 79.68) +- (0.12000000000000455,0.12000000000000455)
			(0.5, 79.10499999999999) +- (0.22500000000000142,0.22500000000000142)
			(1.0, 78.435) +- (0.15500000000000114,0.15500000000000114)
			(1.5, 77.63) +- (0.12999999999999545,0.12999999999999545)
			(2.0, 76.525) +- (0.4750000000000014,0.4750000000000014)
			(2.5, 75.455) +- (0.394999999999996,0.394999999999996)
			(3.0, 74.23) +- (0.23000000000000398,0.23000000000000398)
			(3.5, 72.38) +- (0.7100000000000009,0.7100000000000009)
			(4.0, 71.15) +- (0.6600000000000037,0.6600000000000037)
			(4.5, 69.705) +- (0.394999999999996,0.394999999999996)
			(5.0, 67.245) +- (0.9750000000000014,0.9750000000000014)
			(5.5, 65.305) +- (0.5850000000000009,0.5850000000000009)
			(6.0, 63.545) +- (0.754999999999999,0.754999999999999)
			(6.5, 61.379999999999995) +- (1.2800000000000047,1.2800000000000047)
			(7.0, 58.92) +- (1.6999999999999957,1.6999999999999957)
			(7.5, 56.080000000000005) +- (1.240000000000002,1.240000000000002)
			(8.0, 54.135) +- (0.32499999999999574,0.32499999999999574)
			(8.5, 51.02499999999999) +- (0.495000000000001,0.495000000000001)
			(9.0, 48.615) +- (0.7250000000000014,0.7250000000000014)
			(9.5, 45.495000000000005) +- (0.46499999999999986,0.46499999999999986)
			(10.0, 42.94) +- (0.9299999999999997,0.9299999999999997) };
		\addlegendentry{flip, $p=0.05$}
		\addplot+[line width=1pt,mark=none,color=2CA02C,error bars/.cd,y dir=both, y explicit] coordinates { (0.0, 78.235) +- (0.11499999999999488,0.11499999999999488)
			(0.5, 77.985) +- (0.2749999999999915,0.2749999999999915)
			(1.0, 77.25) +- (0.12000000000000455,0.12000000000000455)
			(1.5, 76.455) +- (0.125,0.125)
			(2.0, 75.655) +- (0.3050000000000068,0.3050000000000068)
			(2.5, 75.16) +- (0.21000000000000085,0.21000000000000085)
			(3.0, 73.925) +- (0.5450000000000017,0.5450000000000017)
			(3.5, 72.775) +- (0.4650000000000034,0.4650000000000034)
			(4.0, 71.285) +- (0.09499999999999886,0.09499999999999886)
			(4.5, 70.22999999999999) +- (0.020000000000003126,0.020000000000003126)
			(5.0, 68.66999999999999) +- (0.7100000000000009,0.7100000000000009)
			(5.5, 66.63) +- (0.7399999999999949,0.7399999999999949)
			(6.0, 65.16499999999999) +- (0.25500000000000256,0.25500000000000256)
			(6.5, 63.485) +- (0.42499999999999716,0.42499999999999716)
			(7.0, 61.59) +- (0.9399999999999942,0.9399999999999942)
			(7.5, 59.055) +- (0.8049999999999997,0.8049999999999997)
			(8.0, 56.815) +- (1.2749999999999986,1.2749999999999986)
			(8.5, 54.335) +- (0.40500000000000114,0.40500000000000114)
			(9.0, 51.3) +- (1.0500000000000007,1.0500000000000007)
			(9.5, 50.115) +- (0.2849999999999966,0.2849999999999966)
			(10.0, 46.33) +- (1.1000000000000014,1.1000000000000014) };
		\addlegendentry{flip, $p=0.1$}

	\end{axis}
\end{tikzpicture}
    \end{subfigure}%
    \\
    \begin{subfigure}[b]{0.32\textwidth}
     \begin{tikzpicture}
	\begin{axis}[
		xlabel=b,
		ylabel= $T^{b}$,
    height=5cm,
    width=6cm,
		legend pos=south west,
		title = \textbf{FCNN FashionMNIST}
	]

		\addplot+[mark=none,color=1F77B4,error bars/.cd,y dir=both, y explicit] coordinates { (2, 0.6597501933574675) +- (4.500150680541992e-06,4.500150680541992e-06)
			(4, 0.652961283922195) +- (0.0001549422740940165,0.0001549422740940165)
			(8, 0.6395815908908841) +- (0.00018242001533497199,0.00018242001533497199)
			(16, 0.6123365759849546) +- (0.0006870031356815409,0.0006870031356815409)
			(32, 0.557473272085189) +- (0.0014204084873199463,0.0014204084873199463)
			(64, 0.447355911135673) +- (0.0030402392148969892,0.0030402392148969892) };
		\addlegendentry{No Reg.}
		\addplot+[mark=none,color=FF7F0E,error bars/.cd,y dir=both, y explicit] coordinates { (2, 0.660745322704315) +- (2.6583671569990752e-05,2.6583671569990752e-05)
			(4, 0.6547406315803521) +- (1.4305114746038239e-05,1.4305114746038239e-05)
			(8, 0.642780900001525) +- (9.316205978399106e-05,9.316205978399106e-05)
			(16, 0.6189983487129205) +- (0.00013065338134748972,0.00013065338134748972)
			(32, 0.5716938078403471) +- (0.0001007616519930199,0.0001007616519930199)
			(64, 0.47888298332691154) +- (0.0001787692308424793,0.0001787692308424793) };
		\addlegendentry{flip, $p=0.1$}
		\addplot+[mark=none,color=2CA02C,error bars/.cd,y dir=both, y explicit] coordinates { (2, 0.6615693569183345) +- (6.347894668551346e-05,6.347894668551346e-05)
			(4, 0.6564982831478116) +- (7.590651512145996e-05,7.590651512145996e-05)
			(8, 0.6462543308734885) +- (9.819865226751157e-05,9.819865226751157e-05)
			(16, 0.625774502754211) +- (2.9385089873990022e-05,2.9385089873990022e-05)
			(32, 0.58544972538948) +- (0.00012835860252402576,0.00012835860252402576)
			(64, 0.5069977939128871) +- (0.0004746615886689898,0.0004746615886689898) };
		\addlegendentry{flip, $p=0.2$}

	\end{axis}
\end{tikzpicture}
    \end{subfigure}%
    \begin{subfigure}[b]{0.32\textwidth}
     \begin{tikzpicture}
	\begin{axis}[
		xlabel=b,
    width=6cm,
    height=5cm,
    legend pos=south west,
		title = \textbf{CNN FashionMNIST},
	]

		\addplot+[mark=none,color=1F77B4,error bars/.cd,y dir=both, y explicit] coordinates { (2, 0.7342766821384425) +- (0.0008961260318754993,0.0008961260318754993)
			(4, 0.719299584627151) +- (0.0010122358798979603,0.0010122358798979603)
			(8, 0.6888613700866695) +- (0.0018235445022584673,0.0018235445022584673)
			(16, 0.633766263723373) +- (0.002435177564621027,0.002435177564621027)
			(32, 0.537660986185073) +- (0.0042911469936370294,0.0042911469936370294)
			(64, 0.3997108638286585) +- (0.0015845596790315108,0.0015845596790315108) };
		\addlegendentry{No Reg.}
		\addplot+[mark=none,color=FF7F0E,error bars/.cd,y dir=both, y explicit] coordinates { (2, 0.7331886589527126) +- (0.00042399764060951917,0.00042399764060951917)
			(4, 0.7155913412570944) +- (0.0008948147296904962,0.0008948147296904962)
			(8, 0.6829218566417691) +- (0.001771181821823009,0.001771181821823009)
			(16, 0.6197429001331325) +- (0.0009249746799465108,0.0009249746799465108)
			(32, 0.517264425754547) +- (0.004239261150359996,0.004239261150359996)
			(64, 0.37031896412372556) +- (0.00531445443630249,0.00531445443630249) };
		\addlegendentry{flip, $p=0.05$}
		\addplot+[mark=none,color=2CA02C,error bars/.cd,y dir=both, y explicit] coordinates { (2, 0.7330339252948761) +- (0.000488132238388006,0.000488132238388006)
			(4, 0.7157599925994871) +- (0.001242220401764027,0.001242220401764027)
			(8, 0.6828210353851311) +- (0.0023161768913270153,0.0023161768913270153)
			(16, 0.6197084188461295) +- (0.003651380538940485,0.003651380538940485)
			(32, 0.5197539925575255) +- (0.004332304000854492,0.004332304000854492)
			(64, 0.39254342019557953) +- (0.0030539482831954956,0.0030539482831954956) };
		\addlegendentry{flip, $p=0.1$}

	\end{axis}
\end{tikzpicture}
    \end{subfigure}%
    \hspace{-0.6cm}
    \begin{subfigure}[b]{0.32\textwidth}
%
%

\begin{tikzpicture}
	\begin{axis}[
    height=5cm,
    width=6cm,
		xlabel=b,
    title = \textbf{CNN CIFAR10},
		legend pos=south west
	]

		\addplot+[mark=none,color=1F77B4,error bars/.cd,y dir=both, y explicit] coordinates { (2, 0.8772832552591954) +- (6.87802879236594e-05,6.87802879236594e-05)
			(4, 0.8656692902247104) +- (0.00012663181623801673,0.00012663181623801673)
			(8, 0.8424547910690304) +- (0.0002309894833773609,0.0002309894833773609)
			(16, 0.7963663935661315) +- (0.0005450741055766111,0.0005450741055766111)
			(32, 0.7063244183858229) +- (0.0009423103681489004,0.0009423103681489004)
			(64, 0.5414370298385613) +- (0.0014076433938104984,0.0014076433938104984) };
		\addlegendentry{No Reg.}
		\addplot+[mark=none,color=FF7F0E,error bars/.cd,y dir=both, y explicit] coordinates { (2, 0.87771651148796) +- (9.149312973022461e-06,9.149312973022461e-06)
			(4, 0.866589576005935) +- (4.44352626799982e-05,4.44352626799982e-05)
			(8, 0.844226568937301) +- (0.00010302662849398514,0.00010302662849398514)
			(16, 0.7999554276466365) +- (0.0002039074897765003,0.0002039074897765003)
			(32, 0.713358551263809) +- (0.0002721846103669878,0.0002721846103669878)
			(64, 0.553226917982101) +- (0.00045600533485395944,0.00045600533485395944) };
		\addlegendentry{flip, $p=0.05$}
		\addplot+[mark=none,color=2CA02C,error bars/.cd,y dir=both, y explicit] coordinates { (2, 0.8781079351902006) +- (1.904368400551526e-05,1.904368400551526e-05)
			(4, 0.8672968745231626) +- (3.248453140253238e-05,3.248453140253238e-05)
			(8, 0.8457360267639156) +- (8.177757263150287e-05,8.177757263150287e-05)
			(16, 0.8028877377510066) +- (0.00010806322097750565,0.00010806322097750565)
			(32, 0.7191092967987056) +- (0.00022172927856450864,0.00022172927856450864)
			(64, 0.5644151568412775) +- (0.00026267766952547955,0.00026267766952547955) };
		\addlegendentry{flip, $p=0.1$}

	\end{axis}
\end{tikzpicture}
    \end{subfigure}%
    \caption{The experiment results for our proposed flip-training. In the top row are the accuracies plotted over bit error rate. In the bottom row are the $T^b$ values plotted over $b$.}
    \label{Fig:plotgrid}
\end{figure*}
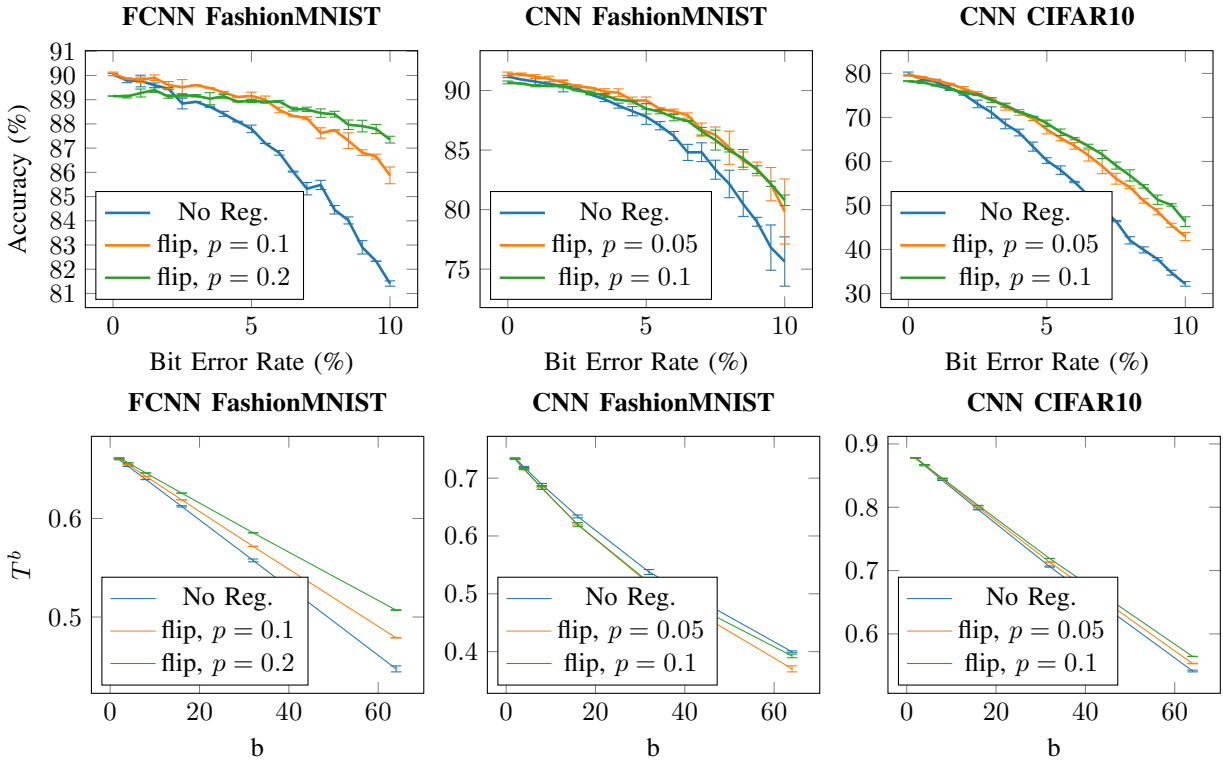

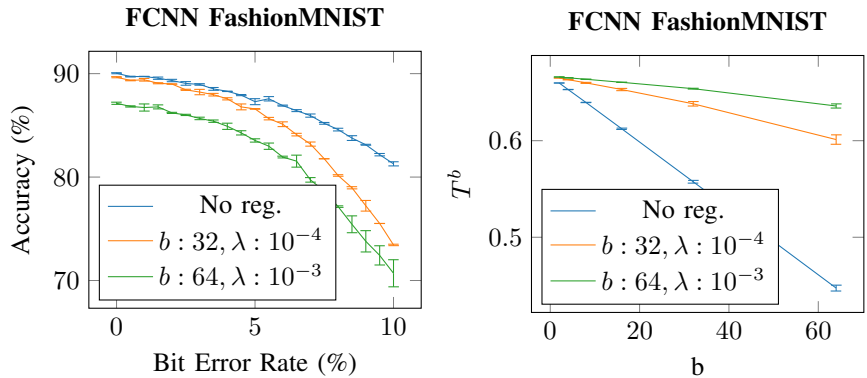
\begin{figure*}[!htb]
    \centering
    \begin{subfigure}[b]{0.32\textwidth}
      \begin{tikzpicture}
	\begin{axis}[
		xlabel=Bit Error Rate (\%),
  	ylabel= Accuracy (\%),
    width=6cm,
    height=5cm,
    legend pos=south west,
		title = \textbf{FCNN FashionMNIST},
	]
	]

		\addplot+[mark=none,color=1F77B4,error bars/.cd,y dir=both, y explicit] coordinates { (0.0, 90.04) +- (0.04999999999999716,0.04999999999999716)
			(0.5, 89.715) +- (0.03499999999999659,0.03499999999999659)
			(1.0, 89.725) +- (0.014999999999993463,0.014999999999993463)
			(1.5, 89.56) +- (0.10999999999999943,0.10999999999999943)
			(2.0, 89.33500000000001) +- (0.11500000000000199,0.11500000000000199)
			(2.5, 89.055) +- (0.17499999999999716,0.17499999999999716)
			(3.0, 88.95500000000001) +- (0.07499999999999574,0.07499999999999574)
			(3.5, 88.525) +- (0.13500000000000512,0.13500000000000512)
			(4.0, 88.29) +- (0.01999999999999602,0.01999999999999602)
			(4.5, 87.92500000000001) +- (0.054999999999999716,0.054999999999999716)
			(5.0, 87.28999999999999) +- (0.28999999999999915,0.28999999999999915)
			(5.5, 87.57) +- (0.21000000000000085,0.21000000000000085)
			(6.0, 86.91) +- (0.05000000000000426,0.05000000000000426)
			(6.5, 86.425) +- (0.09499999999999886,0.09499999999999886)
			(7.0, 85.945) +- (0.15499999999999403,0.15499999999999403)
			(7.5, 85.195) +- (0.08500000000000085,0.08500000000000085)
			(8.0, 84.59) +- (0.12999999999999545,0.12999999999999545)
			(8.5, 83.77000000000001) +- (0.22999999999999687,0.22999999999999687)
			(9.0, 83.125) +- (0.045000000000001705,0.045000000000001705)
			(9.5, 82.16499999999999) +- (0.11500000000000199,0.11500000000000199)
			(10.0, 81.28) +- (0.19999999999999576,0.19999999999999576) };
		\addlegendentry{No reg.}
		\addplot+[mark=none,color=FF7F0E,error bars/.cd,y dir=both, y explicit] coordinates { (0.0, 89.655) +- (0.054999999999999716,0.054999999999999716)
			(0.5, 89.345) +- (0.035000000000003695,0.035000000000003695)
			(1.0, 89.405) +- (0.125,0.125)
			(1.5, 89.07) +- (0.05000000000000426,0.05000000000000426)
			(2.0, 88.995) +- (0.015000000000000568,0.015000000000000568)
			(2.5, 88.435) +- (0.054999999999999716,0.054999999999999716)
			(3.0, 88.22) +- (0.259999999999998,0.259999999999998)
			(3.5, 87.965) +- (0.1250000000000071,0.1250000000000071)
			(4.0, 87.565) +- (0.10499999999999687,0.10499999999999687)
			(4.5, 86.785) +- (0.2850000000000037,0.2850000000000037)
			(5.0, 86.59) +- (0.02999999999999403,0.02999999999999403)
			(5.5, 85.65) +- (0.11999999999999744,0.11999999999999744)
			(6.0, 85.11) +- (0.23000000000000398,0.23000000000000398)
			(6.5, 84.105) +- (0.125,0.125)
			(7.0, 83.19) +- (0.20000000000000284,0.20000000000000284)
			(7.5, 81.77) +- (0.0,0.0)
			(8.0, 80.155) +- (0.07500000000000284,0.07500000000000284)
			(8.5, 78.965) +- (0.10499999999999687,0.10499999999999687)
			(9.0, 77.225) +- (0.5150000000000006,0.5150000000000006)
			(9.5, 75.515) +- (0.0049999999999954525,0.0049999999999954525)
			(10.0, 73.44) +- (0.06999999999999318,0.06999999999999318) };
		\addlegendentry{$b:32, \lambda:10^{-4}$}
		\addplot+[mark=none,color=2CA02C,error bars/.cd,y dir=both, y explicit] coordinates { (0.0, 87.13499999999999) +- (0.10499999999999687,0.10499999999999687)
			(0.5, 86.83) +- (0.060000000000002274,0.060000000000002274)
			(1.0, 86.725) +- (0.34500000000000597,0.34500000000000597)
			(1.5, 86.785) +- (0.19500000000000028,0.19500000000000028)
			(2.0, 86.20500000000001) +- (0.054999999999999716,0.054999999999999716)
			(2.5, 86.01) +- (0.060000000000002274,0.060000000000002274)
			(3.0, 85.67) +- (0.12000000000000455,0.12000000000000455)
			(3.5, 85.405) +- (0.11500000000000199,0.11500000000000199)
			(4.0, 84.91) +- (0.29999999999999716,0.29999999999999716)
			(4.5, 84.275) +- (0.2050000000000054,0.2050000000000054)
			(5.0, 83.53999999999999) +- (0.1600000000000037,0.1600000000000037)
			(5.5, 82.98499999999999) +- (0.3049999999999997,0.3049999999999997)
			(6.0, 81.935) +- (0.08500000000000796,0.08500000000000796)
			(6.5, 81.53999999999999) +- (0.5800000000000054,0.5800000000000054)
			(7.0, 79.73) +- (0.21999999999999886,0.21999999999999886)
			(7.5, 78.64) +- (0.35999999999999943,0.35999999999999943)
			(8.0, 77.14500000000001) +- (0.07499999999999574,0.07499999999999574)
			(8.5, 75.435) +- (0.8149999999999977,0.8149999999999977)
			(9.0, 73.78999999999999) +- (1.029999999999994,1.029999999999994)
			(9.5, 72.43) +- (0.9299999999999997,0.9299999999999997)
			(10.0, 70.69999999999999) +- (1.3099999999999952,1.3099999999999952) };
		\addlegendentry{$b:64, \lambda:10^{-3}$}

	\end{axis}
\end{tikzpicture}
    \end{subfigure}%
    \begin{subfigure}[b]{0.32\textwidth}
      \begin{tikzpicture}
	\begin{axis}[
		xlabel=b,
		ylabel= $T^{b}$,
    height=5cm,
    width=6cm,
		legend pos=south west,
		title = \textbf{FCNN FashionMNIST}
	]	
		\addplot+[mark=none,color=1F77B4,error bars/.cd,y dir=both, y explicit] coordinates { (2, 0.6597501933574675) +- (4.500150680541992e-06,4.500150680541992e-06)
			(4, 0.652961283922195) +- (0.0001549422740940165,0.0001549422740940165)
			(8, 0.6395815908908841) +- (0.00018242001533497199,0.00018242001533497199)
			(16, 0.6123365759849546) +- (0.0006870031356815409,0.0006870031356815409)
			(32, 0.557473272085189) +- (0.0014204084873199463,0.0014204084873199463)
			(64, 0.447355911135673) +- (0.0030402392148969892,0.0030402392148969892) }; 
			\addlegendentry{No reg.} 
		\addplot+[mark=none,color=FF7F0E,error bars/.cd,y dir=both, y explicit] coordinates { (2, 0.6649628877639766) +- (0.0001181960105895441,0.0001181960105895441)
			(4, 0.6632128655910485) +- (0.0002647936344145063,0.0002647936344145063)
			(8, 0.659879028797149) +- (0.0005375742912289705,0.0005375742912289705)
			(16, 0.6529803574085231) +- (0.0011648237705230158,0.0011648237705230158)
			(32, 0.638202399015426) +- (0.0023247897624970038,0.0023247897624970038)
			(64, 0.6011743545532225) +- (0.004878938198089489,0.004878938198089489) }; 
		
		\addlegendentry{$b:32, \lambda:10^{-4}$} 
		\addplot+[mark=none,color=2CA02C,error bars/.cd,y dir=both, y explicit] coordinates { (2, 0.6659927666187281) +- (6.0111284255981445e-05,6.0111284255981445e-05)
			(4, 0.6651248335838316) +- (5.4001808166503906e-05,5.4001808166503906e-05)
			(8, 0.6636174917221065) +- (6.264448165854697e-05,6.264448165854697e-05)
			(16, 0.6604556441307061) +- (0.00014752149581903629,0.00014752149581903629)
			(32, 0.6538803279399871) +- (0.00042530894279502185,0.00042530894279502185)
			(64, 0.635954827070236) +- (0.0021584331989289995,0.0021584331989289995) }; 
		
		\addlegendentry{$b:64, \lambda:10^{-3}$}

	\end{axis}
\end{tikzpicture}
    \end{subfigure}%
    \caption{The experiment results for the direct regularization. }
    \label{Fig:T-train}
\end{figure*}

In this section we present our experiment results.
We evaluate fully connected neural networks (FCNNs) and convolutional neural networks (CNNs) in the configurations shown in Table \ref{tab:expsetting} FashionMNIST and CIFAR10 (see Table \ref{tab:datasets}). In all experiments we run the Adam optimizer for $100$ epochs for FashionMNIST and $250$ epochs for CIFAR10 to minimize the cross entropy loss. We use a batch size of $128$ and an initial learning rate of $10^{-3}$. To stabilize training we exponentially decrease the learning rate every $25$ epochs by $50$ percent. All experiments are repeated $5$ times. 
First, we plot the accuracy over bit error rate for NNs trained with straight-through gradient approximation in the top row of Figure \ref{Fig:plotgrid}.
Then, we show the correlation between $T^{b}$ and accuracy over bit error rate in the bottom row of Figure \ref{Fig:plotgrid}. Finally, we evaluate the impact of our proposed direct regularizer on the error tolerance in Figure \ref{Fig:T-train}.

We notice that flip regularization improves the accuracy when bit errors are introduced. This effect is stronger for FCNNs. 
Moreover we see that we can trade a high accuracy at small error rates with a high accuracy at larger error rates. However we do not observe an overfitting to a particular bit error probability.
Second, in the case of FCNNs trained on FashionMNIST and CNNs trained on CIFAR10, we observe that the accuracy over different error rates indeed correlates with $T^b$. 
For the case of CNNs on FashionMNIST, a correlation cannot be observed. 
Overall we see that CNNs are more brittle than FCNNs. This is likely due to the weight-sharing in CNNs, 
where a flip in a convolution filter has effects at every position in the feature map. 
This difference is also reflected in the $T^b$ values. 
In conclusion we see that the $T^b$ measure is better suited for fully connected networks.

Figure \ref{Fig:T-train} depicts the results for the direct regularization training introduced in Section~\ref{subsec:directreg}.
We observe that this training method does not increase the accuracy over error rate, although the $T^b$ values are high. 
Instead regularizing the training objective this way decreases accuracy at any error rate.
Similar curve progressions can be observed for other hyperparameter settings and the CIFAR10 dataset. For smaller regularization scalings $\lambda$, the observed curves approach the unregularized curves, however we never obtain higher accuracies at any error rate. We conclude that our regularizer is currently unusable: While it effectively increases $T^b$, it does so by sacrificing accuracy thereby rendering the resulting models useless.

%
%

\section{Related Work}
\label{sec:RelatedWork}

Deep Nets offer remarkable performance in state of the art image classification tasks, but require immense computation power during training and during inference.
Thus, a natural research question in this context is to ask, whether we can reduce the computation and memory requirements of Deep Nets without hurting its performance.
A common approach to reduce both, memory and computation demands, is to quantize the weights of an already trained network after training is completed.
This way, weights can be stored using fewer bits and fixed point arithmetics can be exploited during inference.
However, this post-processing step usually degrades the classification performance, which leads to sub-optimal performance \cite{lin/etal/2016,Courbariaux/etal/2014,Han/etal/2015}.
More evolved approaches incorporate quantization directly into the training, so that nets can retain their accuracy. Here two approaches exists:

The first approach aims to perform all operations during training (including gradient computation) with fixed point arithmetics.
This way, the network is always restricted to fixed point values and efficient accelerations of the training is enabled by the means of FPGAs and GPUs.
However, such an approach must guarantee a certain numerical precision so that gradient updates are still meaningful and it has to make sure that gradient estimates during training are still unbiased \cite{Gupta/etal/2015}.
The second approach only quantizes the network during the forward pass, but performs all gradient operations with full floating point precision.
This way, gradient updates can be performed with full floating point precision, while the networks performance is based on its fixed-point weights.
In its most extreme version this approach restricts all intermediate calculations and weights to only two values, e.g. `+1' and `-1'\cite{Rastegari/etal/2016, Hubara/etal/2016}.
Since gradient computations are still performed with floating-point precision, this approach still enables regular optimization with stochastic gradient descent and possible regularization, e.g. to enhance the robustness of neural networks against bit errors.

Nonetheless, the error tolerance training of BNNs for low-power memories has not received much attention yet.
Recent work related to BNNs on NVMs focuses more on the realization of the low-power in-memory processing of BNNs than on the error tolerance training aspect. 
E.g., in \cite{hirtzlin2019implementing} Hirtzlin et al. propose to use Spin Torque Magnetoresistive RAM (ST-MRAM) for the in-memory implementation of NNs. 
In their work they highlight the inherent bit error tolerance of BNNs that were trained without robustness training. 
They show that half the energy can be saved without accuracy loss when using a low power setting to write to the memory cells.    

\section{Conclusion}
\label{sec:Conclusion}

In this work, we improved the state-of-the-art bit error tolerance training for BNNs and evaluated a metric that aims to explain the achieved tolerance.
We were able to eliminate the NNs' overfitting to the error rates by employing a special flip-operator with straight-through gradient approximation in the gradient computation.
For BNNs trained with our flip-regularization, we evaluated the robustness metric and found that it correlates with accuracy over error rate for all FCNNs tested.
CNNs trained on FashionMNIST with our improved flip-regularization do not show a high robustness value $T^b$; we hypothesize this is because of the weight sharing property of CNNs and their more complex layer structure.

We also tried to optimize the NNs with respect to the robustness metric $T^b$.
Although we can achieve high $T^b$ values, it does not lead to a better accuracy over error rate.
We think that this is mainly due to regularizing each neuron equally in our model.
This ignores the effect that a highly robust second layer can compensate low robustness of the first layer. 

In the future, we aim to improve the robustness metric $T^b$, so that error tolerance is better described by it.
In the flip-regularization, only the error rate can be configured, therefore we also aim to improve our direct regularization method, so that the error tolerance can be more finely tuned, e.g. with a configurable trade-off between accuracy and bit error tolerance.



\bibliography{main}
\bibliographystyle{ieeetr}
\appendix




\begin{proof}[Proof for Theorem~\ref{thm:tolerateflip}]
	At first we consider the $n$-th neuron and assume that it is not in the first layer.
	Let $u,v$ be a position for the convolution result.
	As described in \eqref{eq:Tofy}, we know that $T_{i,n,u,v} = | h_{i,n,u,v} - s_n -\frac{1}{2}|$. 
	For improved readability, we write $h$ for $h_{i,n,u,v}$ and $s$ for $s_n$. 
	By construction of the activation function, the output of the neuron at position $u,v$ is $+1$ for $h - s-\frac{1}{2}>0$ and $-1$ for $h - s-\frac{1}{2}<0$ since $h$ and $s$ are assumed to be integer values.
	Furthermore, the case $h - s-\frac{1}{2}=0$ does not occur. 
	If $T_{i,n,u,v} \geq b$ then either $h - s-\frac{1}{2}\geq b$ or $h - s-\frac{1}{2}\leq -b$.
	\par
	In the first case we have  $h - s-\frac{1}{2}\geq b\geq 0$ and the output at $u,v$ is $+1$.
	We denote by $\tilde{y}$ the value of $h$ after the bitflips of up to $\lfloor \frac{b}{2} \rfloor$ weights.
	By definition, $h$ is a weighted sum where each summand is one input of the neuron multiplied with one weight.
	Since each summand is in $\{\pm 1\}$, changing one sign changes $h$ by 2.
	Therefore $\tilde{h}$ can differ by up to $2 \cdot \lfloor \frac{b}{2}\rfloor$ from $h$ and is in $[h-\lfloor b \rfloor, h+\lfloor b \rfloor]$. 
	For $\tilde{h}$ we still have  $\tilde{h}-s-\frac{1}{2}\geq b - \lfloor {b} \rfloor \geq 0$ which causes an output of $+1$ at $u,v$.
	\par
	The second case is proven analogously:
	We have $h-s-\frac{1}{2}\leq -b\leq 0$ and the output of the neuron at $u,v$ is $-1$.
	Changing the sign of $\lfloor \frac{b}{2} \rfloor$ summands of $h$ can increase the value of $h$ by up to $\lfloor {b} \rfloor$.
	After at most $\lfloor \frac{b}{2} \rfloor$ bitflips, we obtain a new value $\tilde{h} \in [h-\lfloor b \rfloor, h+\lfloor b \rfloor]$.
	We have $\tilde{h}-s-\frac{1}{2}\leq -b + \lfloor {b} \rfloor \leq 0$ and the output of this neuron at $u,v$ is still $-1$.
	\par
	If the $n$-th neuron is in the first layer we have $T_{i,n,u,v} = \tfrac{|h - s - \tfrac{1}{2}|}{Z}$ by \eqref{eq:Tofyfirst}.
	Since the summands of $h$ are in $[-Z,Z]$, the value of $h$ can change by up to $2\cdot Z$ per bitflip.
	Thus, after $\lfloor\frac{b}{2}\rfloor$ bitflips the value of $\tilde{h}$ is in the interval $[h- Z\cdot \lfloor b \rfloor, h+Z \cdot \lfloor b\rfloor]$.
	But the proof still works as above since $|h-s-\frac{1}{2}| \geq Z \cdot b$ by assumption.
\end{proof}

\end{document}